\documentclass[runningheads]{llncs}

\usepackage{eccv}

\usepackage{eccvabbrv}

\usepackage{graphicx}
\usepackage{booktabs}
\usepackage{courier}
\usepackage[flushleft]{threeparttable}
\usepackage[accsupp]{axessibility}  
\usepackage{subfloat}

\usepackage{hyperref}

\usepackage{orcidlink}

\begin{document}

\title{Relation DETR: Exploring Explicit Position Relation Prior for Object Detection}

\titlerunning{Relation DETR}

\author{Xiuquan Hou\inst{1}\orcidlink{0009-0000-2547-4921} \and
  Meiqin Liu\inst{1,2,}\thanks{Corresponding author}\orcidlink{0000-0003-0693-6574} \and
  Senlin Zhang\inst{2}\orcidlink{0000-0001-5117-3110}
  Ping Wei\inst{1}\orcidlink{0000-0002-8535-9527} \and
  Badong Chen\inst{1}\orcidlink{0000-0003-1710-3818} \and
  Xuguang Lan\inst{1}\orcidlink{0000-0002-3422-944X}
  }

\authorrunning{X.~Hou et al.}

\institute{National Key Laboratory of Human-Machine Hybrid Augmented Intelligence, Xi'an Jiaotong University, Xi'an 710049, China \and
College of Electrical Engineering, Zhejiang University, Hangzhou 310027, China\\
  \email{xiuqhou@stu.xjtu.edu.cn},\email{\{liumeiqin,slzhang\}@zju.edu.cn},
  \email{\{pingwei,chenbd,xglan\}@mail.xjtu.edu.cn}
  }

\maketitle

\begin{abstract}
  This paper presents a general scheme for enhancing the convergence and performance of DETR (DEtection TRansformer). We investigate the slow convergence problem in transformers from a new perspective, suggesting that it arises from the self-attention that introduces no structural bias over inputs.
  To address this issue, we explore incorporating position relation prior as attention bias to augment object detection, following the verification of its statistical significance using a proposed quantitative macrosopic correlation (MC) metric.
  Our approach, termed Relation-DETR, introduces an encoder to construct position relation embeddings for progressive attention refinement, which further extends the traditional streaming pipeline of DETR into a contrastive relation pipeline to address the conflicts between non-duplicate predictions and positive supervision.
  Extensive experiments on both generic and task-specific datasets demonstrate the effectiveness of our approach.
  Under the same configurations, Relation-DETR achieves a significant improvement (+2.0\% AP compared to DINO), state-of-the-art performance (51.7\% AP for 1$\times$ and 52.1\% AP for $2\times$ settings), and a remarkably faster convergence speed (over $40\%$ AP with \textbf{only 2} training epochs) than existing DETR detectors on COCO val2017. Moreover, the proposed relation encoder serves as a universal plug-in-and-play component, bringing clear improvements for theoretically any DETR-like methods. Furthermore, we introduce a class-agnostic detection dataset, SA-Det-100k. The experimental results on the dataset illustrate that the proposed explicit position relation achieves a clear improvement of 1.3\% AP, highlighting its potential towards universal object detection. The code and dataset are available at \href{https://github.com/xiuqhou/Relation-DETR}{https://github.com/xiuqhou/Relation-DETR}.
\end{abstract}

\section{Introduction}
\label{sec:intro}
Object detection aims to tackle the problems of bounding box regression and object classification for each object of interest. Recently, DEtection TRansformer (DETR) \cite{carion2020end} has overcome the reliance on handcrafted designs of convolution detectors, achieving an elegant architecture in an end-to-end manner.
Despite exhibiting impressive detection performance on large-scale datasets such as COCO \cite{lin2014microsoft},
their performances are prone to be influenced by dataset scale and suffer from slow convergence.
The root cause of the problem is the conflict between non-duplicate predictions and positive supervision \cite{jia2023detrs}. During the training process, DETR employs the Hungarian algorithm to assign a single positive prediction to each ground-truth for producing unique results. However, this leads to negative predictions dominating the majority of the loss function, causing insufficient positive supervision. Therefore, more samples and iterations are required for convergence. Previous attempts have explored the issue by introducing train-only architectures (\eg query denoising \cite{li2022dn}, multiple groups of queries \cite{chen2023group}, auxiliary queries \cite{jia2023detrs}, collaborative hybrid assignment training \cite{zong2023detrs}) for additional supervision or by incorporating hard mining in loss functions (\eg IA-BCE loss \cite{cai2023align}, position-supervised loss \cite{liu2023detection}). Other works have proposed specific structures for better interaction between queries and feature maps (\eg dynamic anchor query \cite{liu2021dab}, cascade window attention \cite{ye2023cascade}), as well as techniques to focus on high-quality queries (\eg hierarchical filtering \cite{hou2024salience}, dense distinct process \cite{zhang2023dense} and query rank layer \cite{pu2024rank}). Despite these advancements, there has been little exploration of the issue from the perspective of self-attention, which is widely used in the transformer decoders in most DETR detectors.

The effectiveness of self-attention lies in its establishment of a high-dimensional relation representation among sequence embeddings \cite{carion2020end,vaswani2017attention}, which also serves as a key component for modeling relations among different detection feature representations \cite{carion2020end}. However, such relation is an implicit representation since it assumes no structural bias over inputs, in which even position information is also needed to be learned from training data \cite{lin2022survey}. Consequently, the learning process of transformer is data-intensive and slow to converge. This analysis motivates us to introduce task-specific bias for realizing faster convergence and reducing data dependence.

In this paper, we explore enhancing DETR detectors from a novel perspective, namely explicit position relation prior. We first establish a metric for quantifying position relations in images, and analyze the distribution to verify its statistical significance. In this context, we introduce a position relation encoder to model all pairwise interactions between two bounding boxes, employing progressive attention refinement for cross-layer information interaction. To maintain the end-to-end property while providing sufficient positive supervision, we introduce a contrast relation strategy, which leverages both one-to-one and one-to-many matching while emphasizing the influence of position relation on deduplication. The proposed method is named as Relation-DETR.

Compared to previous works, the main feature of Relation-DETR is the integration of explicit position relation. In contrast, prior works focus on implicitly learned attention weights from training data, leading to slow convergence. Intuitively, our proposed position relation can be seen as a plug-in-and-play design beneficial for non-duplication predictions, since it establishes a representation of relative positions among pairs of bounding boxes(similar to IoU in NMS \cite{girshick2015fast}).

We evaluate the performance of Relation-DETR on the most popular object detection dataset, COCO 2017 \cite{lin2014microsoft}, as well as several task-specific datasets \cite{wang2023surface,chen2023eee}. The experimental results demonstrate its superior performance, surpassing previous state-of-the-art DETR detectors with clear margins. More specifically, Relation-DETR exhibits a significantly fast convergence speed. Without bells and whistles, it becomes the the first DETR detector to achieve $40\%$ AP on COCO with only 2 epochs using ResNet50 as the backbone under the $1\times$ training configuration. In addition, the simple architecture design of our position relation encoder maintains a promising transferability. It can be easily extended to other DETR-based methods with only a few modifications to achieve consistent performance improvements. This is in contrast to some existing DETR detectors whose performance is highly dependent from complex matching strategies \cite{hu2024dac} or detection heads \cite{zong2023detrs} developed by convolution-based detectors.

\section{Related Work}
\label{sec:rela}
\subsubsection{Transformer for Object Detection}
\label{sec:trans for obj det}
In practice, the majority of attempts to apply transformer to object detection involve constructing a parallelizable sequence,
either in the feature extractor \cite{li2022exploring} or in the detection body \cite{carion2020end}. Specifically, transformer-based feature extractors generate token sequences based on image patches \cite{liu2021swin,dosovitskiy2020image,li2022exploring}, and extract multi-scale features through aggregating local features \cite{li2022exploring,liu2021swin} or pyramid postprocess \cite{li2022exploring,fang2023eva}. DEtection TRansformer (DETR) proposed by Carion \etal \cite{carion2020end} encodes the extracted features into object queries and decodes them into detected bounding boxes and labels. However, the self-learned attention mechanism increases the requirements for large-scale datasets and training iterations. Many works have explored the slow convergence from the perspective of structured attention (\eg multi-scale deformable attention \cite{zhu2020deformable}, dynamic attention \cite{dai2021dynamic}, cascade window attention \cite{ye2023cascade}), queries with explicit priors (\eg anchor queries \cite{wang2022anchor}, dynamic anchor boxes queries \cite{liu2021dab}, denoising queries \cite{li2022dn}, dense distinct queries \cite{zhang2023dense}), and additional positive supervision (\eg group queries \cite{chen2023group}, hybrid design \cite{jia2023detrs}, mixed matching \cite{cai2023align}). However, even state-of-the-art DETR methods still utilize vanilla multi-head attention in the transformer decoder. And few works have explored the slow convergence from the perspective of implicit priors. This paper aims to address the issue with position relation.

\subsubsection{Relation Network}
\label{sec:rela network}
Rather than processing visual features at pixel levels, patch levels or image levels, relation network captures relation features at instance levels. Existing research on relation networks involves category-based and instance-based approaches. The category-based approaches construct conceptual or statistical relations (\eg co-occurrence probability \cite{krishna2017visual,hao2023relation}) either from relation datasets like Visual Genome \cite{krishna2017visual,jiang2018hybrid,xu2019reasoning} or by adaptively learning from class labels \cite{hao2023relation}. Both of them, however, increase the complexity due to the assignment between instances and categories \cite{xu2019reasoning,hao2023relation,chen2018iterative}. In contrast, instance-based approaches directly construct a fine-grained graph structure given object features as a node set and their relations as a edge set. Therefore, reasoning on the graph during the training process naturally determines the explicit relation weight \cite{hu2018relation}. Typically, the weight denotes the parametric distances between each paired object instances in high-dimensional space, such as appearance similarity \cite{li2020gar}, proposal distance \cite{lin2021core} or even self-attention weight \cite{vaswani2017attention}. Since learning self-attention weight solely from training data without structural bias increases the requirement for dataset scale and iterations, we explore explicit position relation as a prior to reduce the requirement.

\subsubsection{Classification loss for hard mining.}
\label{sec:cls loss for HM}
During object detection training, positive predictions assigned to ground truth are much fewer than negative predictions, often resulting in imbalanced supervision and slow convergence. For classification tasks, Focal Loss \cite{lin2017focal} proposes introducing a weight parameter to focus on hard samples, which is further extended to many variants like generalized focal loss (GFL) \cite{li2020generalized}, vari focal loss (VFL) \cite{li2020generalized}. Moreover, for object detection tasks, using loss with modulation terms based on regression metrics (\eg TOOD \cite{feng2021tood}, IA-BCE \cite{cai2023align}, position-supervised loss \cite{liu2023detection}) further achieves high-quality alignment between classification and regression tasks.

\section{Statistical significance of object position relation}
\label{sec:significance of relation}
Are objects really correlated in object detection tasks? To answer the question, we propose a quantitative macroscopic correlation (MC) metric based on the Pearson Correlation Coefficient (PCC) to measure the position correlation among objects in a single image.
Assume the objects in an image form a node set, and the PCC between each pair of bounding box annotations serves as their corresponding edge weight. We can construct an undirected graph with continuous values. In this end, the macroscopic correlation for each image can be calculated using the graph intensity, formulated as:
\begin{equation}
  MC = \frac{\sum_i\sum_{j:j\ne i}\left|\texttt{Pearson}(\boldsymbol b_i, \boldsymbol b_j)\right|}{N(N-1)}
\end{equation}
where $N$ denotes the number of objects, \ie the number of nodes, $\boldsymbol b=[x, y, w, h]$ denotes the position annotation of bounding boxes in datasets. $MC=1$ only when all objects are fully linearly correlated, whereas $MC=0$ if there is no position correlation between any pair of objects.

We visualize the statistical distribution of MC for datasets across various scenarios, including industrial (ESD \cite{hou2023canet}, CSD \cite{wang2023surface}, MSSD \cite{chen2023eee}), domestic (AI2thor \cite{kolve2017ai2}), urban (Cityscapes \cite{cordts2016cityscapes}) and generic settings (PascalVOC \cite{everingham2010pascal}, COCO \cite{lin2014microsoft}, Object365 \cite{shao2019objects365}, SA-1B \cite{kirillov2023segment}). The datasets cover a wide range of scales, from 0.3K to 11M images.
\begin{figure}[htbp]
  \centering
  \includegraphics{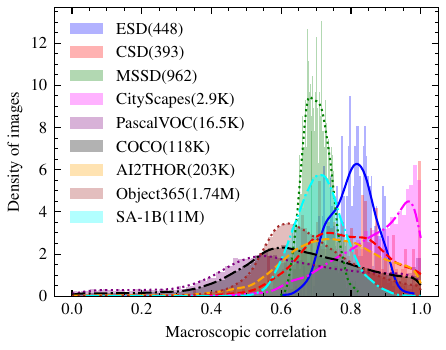}
  \caption{Statistical distribution of macroscopic correlation (MC) on various datasets (normalized for better visualization), and the values in brackets indicate the number of dataset samples.}
  \label{fig:statis mc}
\end{figure}
As shown in \cref{fig:statis mc}, all of these datasets indicate that the distribution of MC is concentrated around high numerical values, with the distribution centers closer to the upper bound. This demonstrates the presence and the statistical significance of object position relation. Specifically, task-specific datasets display more prior knowledge and clearer clustering patterns in the high-dimensional feature space, which thus results in higher MC values than generic datasets like COCO.

\section{Relation-DETR}
\label{sec:relation-detr}
Given the statistical significance of position relation, we propose a state-of-the-art detector, named Relation-DETR, that explores explicit position relation prior to enhance object detection. To address the issue of slow convergence, we present a position relation encoder(\cref{sec:position relation encoder}) for progressive attention refinement(\cref{sec:progressive attention refinement with position relation}). Further in \cref{sec:contrast relation pipeline}, it extends the streaming pipeline of DETR into a contrast pipeline to emphasize the influence of position relation on removing duplication while maintaining sufficient positive supervision for faster convergence.
\subsection{Position relation encoder}
\label{sec:position relation encoder}
Previous research has demonstrated the relation effectiveness for convolution detectors \cite{hu2018relation,lin2021core}. Recently, some DETR methods attempt to construct instance-level relation by indexing from category-level relation using class indices \cite{hao2023relation}. In contrast to these approaches, we directly construct instance-level relation through a simple position encoder, maintaining an end-to-end design for DETR.

We first review the basic pipeline in DETR detectors. Given image features extracted by the backbone, the transformer encoder produces an augmented memory $\mathbf Z\in\mathbb R^{d\times H\times W}$ for further decoding into bounding boxes $\boldsymbol b_i=[x,y,w,h], i=1,\cdots,N$ and class labels $\boldsymbol c_i$ as predictions. Each decoder layer refines the bounding box coordinates iteratively by predicting $\Delta$ \wrt the coordinate from the last decoder layer, known as iterative bounding box refinement \cite{zhu2020deformable}. In addition, predictions from all decoder layers equally participate in loss calculation to compute auxiliary decoding losses \cite{carion2020end}.

Under the aforementioned detection framework, our position relation encoder represents the high-dimensional relation embedding as an explicit prior for self-attention in the transformer. This embedding is calculated based on the predicted bounding boxes (denoted as $\mathbf b=[x, y, w, h]$) from each decoder layer. To ensure that the relation is invariant to translation and scale transformations, we encode it based on normalized relative geometry features:
\begin{equation}
  \boldsymbol e(\boldsymbol b_i, \boldsymbol b_j)=\left[\log\left(\frac{|x_i-x_j|}{w_i}\!+\!1\right)\!, \log\left(\frac{|y_i-y_j|}{h_i}\!+\!1\right)\!, \log\left(\frac{w_i}{w_j}\right)\!, \log\left(\frac{h_i}{h_j}\right)\right]
\end{equation}
Unlike \cite{hu2018relation}, our position relation is unbiased, as $\boldsymbol e(\boldsymbol b_i,\boldsymbol b_j)=0$ when $i=j$. The relation matrix $\mathbf E\in\mathbb R^{N\times N\times 4}$ (with $\mathbf E(i,j)=\boldsymbol e(\boldsymbol b_i,\boldsymbol b_j)$) is further transformed into high-dimensional embeddings through sine-cosine encoding \cite{vaswani2017attention}.
\begin{align}
  \texttt{Embed}(\mathbf{E},2k)=\sin\left(s\mathbf{E}/T^{2k/d_\text{re}}\right) \\
  \texttt{Embed}(\mathbf{E},2k+1)=\cos\left(s\mathbf{E}/T^{2k/d_\text{re}}\right)
\end{align}
where the shape of relation embedding is $N\times N\times 4d_\text{re}$, and $T$, $d_\text{re}$, $s$ are encoding parameters. Finally, the embedding undergoes a linear transformation to obtain $M$ scalar weights, where $M$ denotes the number of attention heads.
\begin{equation}\label{eq:relation embedding}
  \texttt{Rel}(\boldsymbol b,\boldsymbol b)=\max{(\epsilon, \mathbf W\texttt{Embed}(\boldsymbol b,\boldsymbol b)+\mathbf B)}
\end{equation}
where $\epsilon$ makes sure a positive value for relation to avoid gradient vanishing after $\texttt{exp}$ when integrated into self-attention, and $\texttt{Rel}(\boldsymbol b, \boldsymbol b)\in\mathbb R^{N\times N\times M}$.

\subsection{Progressive attention refinement with position relation}
\label{sec:progressive attention refinement with position relation}
The iterative box refinement proposed by Deformable-DETR \cite{zhu2020deformable} has shown the effectiveness for high-quality bounding box regression. Following the motivation, we propose a progressive attention refinement method to introduce the position relation into the streaming pipeline of DETR. Specifically, the relation of layer-$i$ is determined by bounding boxes of both layer-$i-1$ and layer-$i$, which is further integrated into self-attention for producing the bounding boxes in layer-$(i+1)$.
\begin{align}
  \texttt{Attn}_\texttt{Self}(\mathbf Q^l) & =\texttt{Softmax}\!\left(\textcolor{red}{\texttt{Rel}(\boldsymbol b^{l-1},\boldsymbol b^{l})} + \frac{\texttt{Que}(\mathbf Q^l)\texttt{Key}(\mathbf Q^l)^\top}{\sqrt{d_{model}}}\right)\!\texttt{Val}(\mathbf Q^l) \\
  \mathbf Q^{l+1}                          & =\texttt{FFN}\left(\mathbf Q^l+\texttt{Attn}_\texttt{cross}\left(\texttt{Attn}_\texttt{self}(\mathbf Q^l),\texttt{Key}(\mathbf{Z}),\texttt{Val}(\mathbf{Z})\right)\right)                                          \\
  \boldsymbol b^{l+1}                      & =\texttt{MLP}(\mathbf Q^{l+1}), \boldsymbol c^{l+1}=\texttt{Linear}(\mathbf Q^{l+1})
\end{align}
where $\mathbf Q^l$ denotes the queries in the $l$-th decoder layer in the DETR transformer, $\mathbf{Z}$ is the memory, \ie, the enhanced image features from the transformer encoder.

\begin{figure}[htbp]
  \centering
  \includegraphics[width=0.49\textwidth]{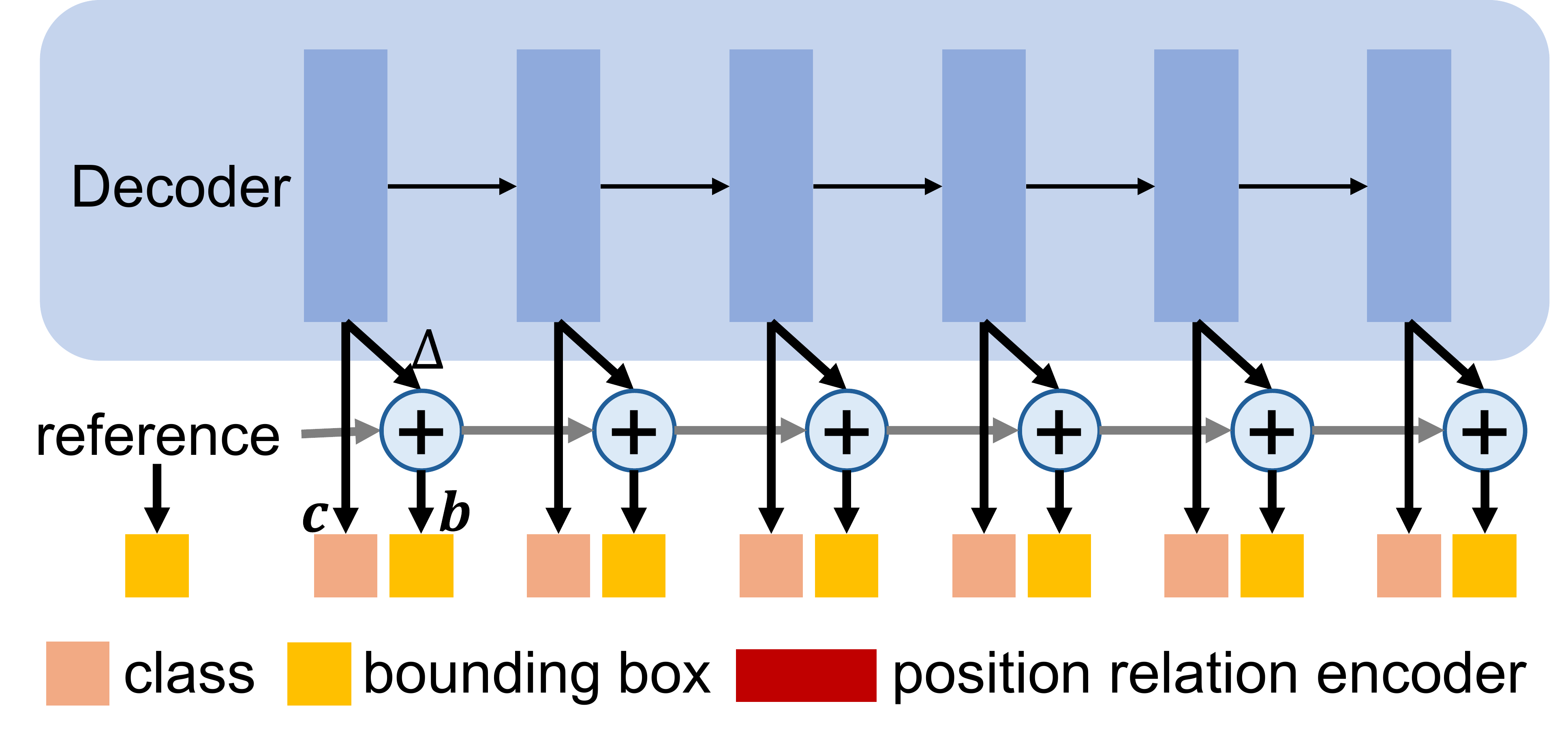}
  \includegraphics[width=0.49\textwidth]{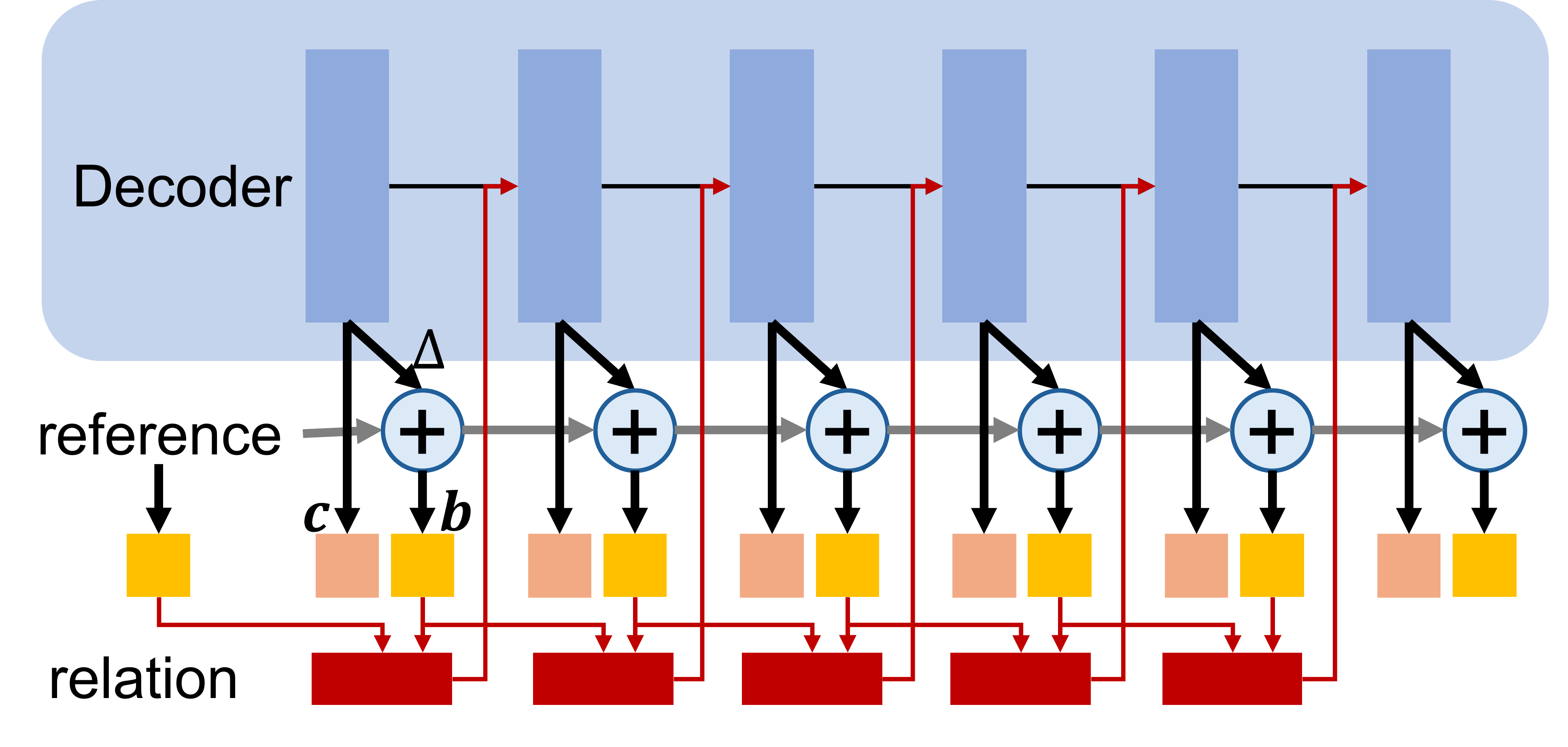}
  \caption{Comparison of transformer decoder in Deformable-DETR(left) and Relation-DETR(right).}
  \label{fig:decoder comparison}
\end{figure}

The only difference between our method and the existing DETR decoder is marked in \textcolor{red}{red}. As depicted in \cref{fig:decoder comparison}, the requisite addition involves the introduction of a lateral branch for the computation of position relation. Therefore, our position relation and the progressive attention refinement are straightforward, allowing for plug-in-and-play integration with self attention in existing DETR detectors to achieve consistent performance improvements (see \cref{tab:transfer exp}).

\subsection{Contrast relation pipeline}
\label{sec:contrast relation pipeline}
Rethinking the mechanism of existing duplication removal methods (including NMS \cite{girshick2015fast}, Soft-NMS \cite{bodla2017soft}, fast-NMS \cite{bolya2019yolact}, Adaptive-NMS \cite{liu2019adaptive}), these processes heavily rely on IoU (intersection over Union), which, to some extent, signifies the position relation between bounding boxes. Therefore, we may hypothesize that integrating the position relation among queries in self-attention contributes to non-duplicated predictions in object detection, akin to \cite{hu2024dac}.

\begin{figure}[htbp]
  \centering
  \includegraphics[width=0.72\textwidth]{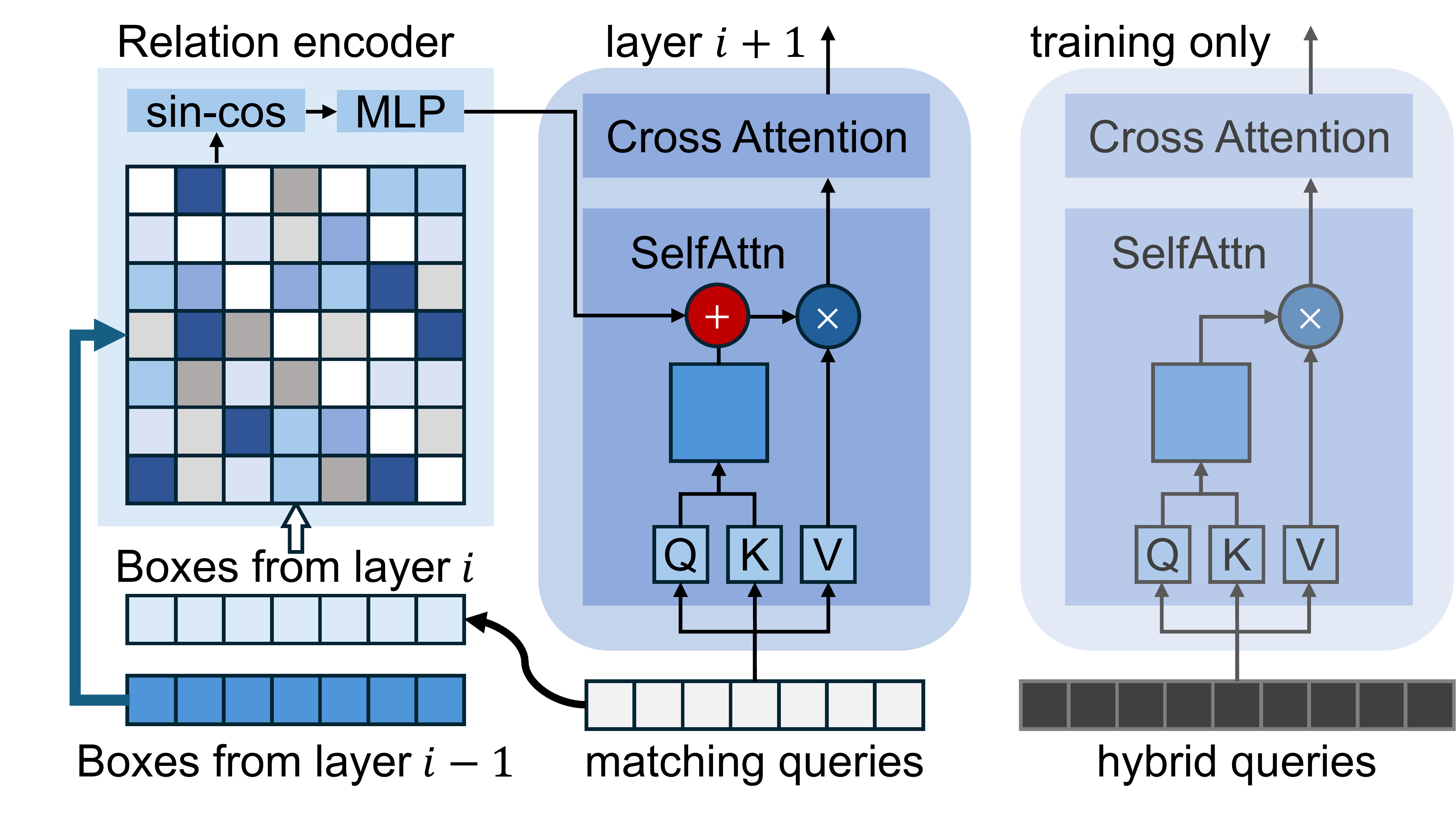}
  \caption{Detailed illustration of the proposed contrast relation pipeline.}
  \label{fig:contrast relation pipeline}
\end{figure}

The conflicts between non-duplicate predictions and sufficient positive supervision arise from the streaming pipeline of DETR, which must navigate between one-to-one matching and one-to-many matching. To overcome this limitation, we extend it to a contrast pipeline based on the proposed position relation. Specifically, we construct two parallel sets of queries, \ie matching queries $\mathbf Q_m$ and hybrid queries $\mathbf Q_h$. Both are input into the transformer decoder but undergo distinct processing. The matching queries are processed with self-attention incorporating position relation to produce non-duplicated predictions:
\begin{align}
  \texttt{Attn}_\texttt{Self}(\mathbf Q_\text{m}^l) & =\texttt{Softmax}\left(\texttt{Rel}(\boldsymbol b^{l-1},\boldsymbol b^{l}) + \frac{\texttt{Que}(\mathbf Q_\text{m})\texttt{Key}(\mathbf Q_\text{m})^\top}{\sqrt{d_{model}}}\right)\texttt{Val}(\mathbf Q_\text{m}) \\
  \texttt{Attn}_\texttt{Self}(\mathbf Q_\text{h}^l) & =\texttt{Softmax}\left(\frac{\texttt{Que}(\mathbf Q_\text{h})\texttt{Key}(\mathbf Q_\text{h})^\top}{\sqrt{d_{model}}}\right)\texttt{Val}(\mathbf Q_\text{h})
\end{align}
while the hybrid queries are decoded by the same decoder but skip the calculation of position relation to explore more potential candidates. Their corresponding predictions are denoted as $\boldsymbol p_m^l=(\boldsymbol b_m^l, \boldsymbol c_m^l)$ and $\boldsymbol p_h^l=(\boldsymbol b_h^l, \boldsymbol c_h^l)$, respectively. Details of the contrast relation pipeline are illustrated in \cref{fig:contrast relation pipeline}.

Assuming $\boldsymbol g$ denotes the ground truth annotations, for $\boldsymbol p_m$, we employ a one-to-one matching scheme to emphasize the non-duplicate property, with loss calculation similar to the original DETR approach \cite{carion2020end}:
\begin{equation}
  \mathcal L_m(\boldsymbol p_m,\boldsymbol g)=\sum_{l=1}^L\mathcal L_\text{Hungarian}(\boldsymbol p_m^l,\boldsymbol g)
\end{equation}
While for $\boldsymbol p_h$, a one-to-many matching scheme is employed to form more potential positive candidates. We simply follow H-DETR \cite{jia2023detrs} and repeat the ground truth $K$ times, denoted as $\tilde{\boldsymbol g}=\{\boldsymbol g^1,\boldsymbol g^2,\cdots,\boldsymbol g^K\}$, for loss calculation:
\begin{equation}
  \mathcal L_h(\boldsymbol p_h,\boldsymbol g)=\sum_{l=1}^L\mathcal L_\text{Hungarian}(\boldsymbol p_h^l,\tilde{\boldsymbol g})
\end{equation}
where $\mathcal L_\text{Hungarian}$ denotes the Hungarian loss, and $L$ denotes the number of decoder layers. It is worth noting that the hybrid queries are only involved during training, thus incurring no extra computational burden for inference.

\section{Experimental Results and Discussion}
\label{sec:experi}
\subsection{Setup}
\label{sec:setup}
For a comprehensive evaluation, we conduct experiments on both an object detection benchmark, COCO 2017 \cite{lin2014microsoft}, and two task-specific datasets, CSD \cite{wang2023surface} and MSSD \cite{chen2023eee}.
Detection performance is measured using the standard Average Precision (AP) \cite{lin2014microsoft}.
We train our model on NVIDIA A800 GPU (80GB) and RTX 3090 GPU (24GB) using the AdamW optimizer with an initial learning rate of $1\times 10^{-4}$ and weight decay of $1\times 10^{-4}$. The Relation-DETR is implemented based on the backbone ResNet-50 \cite{he2016deep} and Swin-L \cite{liu2021swin} pretrained from ImageNet \cite{deng2009imagenet}, which are finetuned with a learning rate $1\times 10^{-5}$ during training. The learning rate is reduced by a factor of 0.1 at later stages. The parameters of the position relation encoder, $T$, $d_\text{re}$, $s$ are empirically chosen as $10000$, $16$, $100$, respectively. The hybrid training configurations follow DINO \cite{zhang2022dino} and H-DETR \cite{jia2023detrs}, \ie, $N_m=900$, $N_h=1500$, $k=6$. We adopt the VariFocal loss \cite{zhang2021varifocalnet} for training Relation-DETR. The training batch size is 10 for COCO 2017 and 2 for task-specific datasets. Before fed into detectors, images undergo the same augmentations (random resize, crop and flip) as other DETR detectors.

\subsection{Comparison with state-of-the-art methods}
\label{sec:comparison with state-of-the-art methods}
\subsubsection{Comparison on COCO 2017.}

\cref{tab:exp coco2017} presents the detection performance on COCO \texttt{val} 2017. Compared to other state-of-the-art DETR methods, our approach converges much faster and demonstrates significant improvements of 1.0\% AP, 1.0\% AP$_{50}$ and 0.6\% AP$_{75}$, respectively, suppressing the second best DDQ-DETR \cite{zhang2023dense} with clear margins. Specifically, Relation-DETR achieves 51.7\% AP using only 12 epochs with the ResNet-50 backbone and even outperforms DINO \cite{zhang2022dino} (51.2\% AP) with 36 epochs (3$\times$ faster). More importantly, in contrast to DDQ-DETR \cite{zhang2023dense} and $\mathcal{C}$o-DETR \cite{zong2023detrs} that leverages NMS in the decoder or postprocess for improving precision, our Relation-DETR maintains an end-to-end pipeline, ensuring promising extensibility. When integrated with Swin-L backbone, Relation-DETR outperforms all counterparts, achieving the best 57.8\% AP with a 0.5\% AP improvement, showcasing its excellent scalability for larger model capacity.

\begin{table}[htbp]
  \caption{Comparison with state-of-the-art methods on COCO \texttt{val}2017 using ResNet-50(IN-1K) backbone. The $*$ means that we re-implement the methods and report the corresponding results.}
  \label{tab:exp coco2017}
  \begin{tabular}{@{}lccccccccc@{}}
    \toprule
    Method                                       & Backbone  & Epochs & AP@50:95      & AP@50         & AP@75         & AP$_S$        & AP$_M$        & AP$_L$        \\ \midrule
    Anchor-DETR \cite{wang2022anchor}            & ResNet-50 & 50     & 42.1          & 63.1          & 44.9          & 22.3          & 46.2          & 60.0          \\
    Def-DETR$^*$ \cite{zhu2020deformable}        & ResNet-50 & 12     & 45.4          & 65.0          & 49.1          & 27.2          & 49.6          & 61.0          \\
    DAB-Def-DETR$^*$ \cite{liu2021dab}           & ResNet-50 & 12     & 46.3          & 65.9          & 50.4          & 29.3          & 50.5          & 61.9          \\
    DN-Def-DETR$^*$ \cite{li2022dn}              & ResNet-50 & 12     & 47.2          & 65.3          & 51.3          & 30.2          & 50.6          & 62.4          \\
    DINO$^*$ \cite{zhang2022dino}                & ResNet-50 & 12     & 49.9          & 67.4          & 54.5          & 33.9          & 53.5          & 63.8          \\
    Group-DETR \cite{chen2023group}              & ResNet-50 & 12     & 49.8          &               &               & 32.4          & 53.0          & 64.2          \\
    $\mathcal{H}$-Def-DETR \cite{jia2023detrs}   & ResNet-50 & 12     & 48.7          & 66.4          & 52.9          & 31.2          & 51.5          & 63.5          \\
    Cascade-DETR \cite{ye2023cascade}            & ResNet-50 & 12     & 49.7          & 67.1          & 54.1          & 32.4          & 53.5          & 65.1          \\
    $\mathcal{C}$o-Def-DETR \cite{zong2023detrs} & ResNet-50 & 12     & 49.5          & 67.6          & 54.3          & 32.4          & 52.7          & 63.7          \\
    Align-DETR \cite{cai2023align}               & ResNet-50 & 12     & 50.2          & 67.8          & 54.4          & 32.9          & 53.3          & 65.0          \\
    Stable-DINO \cite{liu2023detection}          & ResNet-50 & 12     & 50.4          & 67.4          & 55.0          & 32.9          & 54.0          & 65.5          \\
    DAC-DETR \cite{hu2024dac}                    & ResNet-50 & 12     & 50.0          & 67.6          & 54.7          &               &               &               \\
    Salience-DETR$^*$ \cite{hou2024salience}       & ResNet-50   & 12     & 50.0          & 67.7         & 54.2           & 33.3          & 54.4          & 64.4          \\
    Rank-DETR \cite{pu2024rank}                  & ResNet-50 & 12     & 50.4          & 67.9          & 55.2          & 33.6          & 53.8          & 64.2          \\
    MS-DETR \cite{zhao2024ms}                    & ResNet-50 & 12     & 50.3          & 67.4          & 55.1          & 32.7          & 54.0          & 64.6          \\
    DDQ-DETR \cite{zhang2023dense}               & ResNet-50 & 12     & 50.7          & 68.1          & 55.7          &               &               &               \\
    Relation-DETR                                & ResNet-50 & 12     & \textbf{51.7} & \textbf{69.1} & \textbf{56.3} & \textbf{36.1} & \textbf{55.6} & \textbf{66.1} \\
    \midrule
    DINO \cite{zhang2022dino}                    & ResNet-50 & 24     & 50.4          & 68.3          & 54.8          & 33.3          & 53.7          & 64.8          \\
    DINO \cite{zhang2022dino}                    & ResNet-50 & 36     & 51.2          & 69.0          & 55.8          & 35.0          & 54.3          & 65.3          \\
    $\mathcal{H}$-Def-DETR \cite{jia2023detrs}   & ResNet-50 & 36     & 50.0          & 68.3          & 54.4          & 32.9          & 52.7          & 65.3          \\
    Salience-DETR$^*$ \cite{hou2024salience}         & ResNet-50 & 24     & 51.2          & 68.9          & 55.7          & 33.9          & 55.5          & 65.6          \\
    DDQ-DETR \cite{zhang2023dense}               & ResNet-50 & 24     & 52.0          & 69.5          & \textbf{57.2} & 35.2          & 54.9          & 65.9          \\
    Relation-DETR                                & ResNet-50 & 24     & \textbf{52.1} & \textbf{69.7} & 56.6          & \textbf{36.1} & \textbf{56.0} & \textbf{66.5} \\
    \bottomrule
  \end{tabular}
\end{table}

\begin{table}[htbp]
  \setlength{\tabcolsep}{2.2pt}
  \centering
  \caption{Comparison with state-of-the-art methods on COCO \texttt{val}2017 using Swin-L(IN-22K) as the backbone.}
  \begin{tabular}{@{}lccccccccc@{}}
    \toprule
    Method                                     & Backbone & Epochs & AP@50:95      & AP@50         & AP@75         & AP$_S$        & AP$_M$        & AP$_L$        \\ \midrule
    DINO \cite{zhang2022dino}                  & Swin-L   & 12     & 56.8          & 75.4          & 62.3          & 41.1          & 60.6          & 73.5          \\
    $\mathcal{H}$-Def-DETR \cite{jia2023detrs} & Swin-L   & 12     & 55.9          & 75.2          & 61.0          & 39.1          & 59.9          & 72.2          \\
    Salience-DETR$^*$ \cite{hou2024salience}       & Swin-L   & 12     & 56.5          & 75.0          & 61.5          & 40.2          & 61.2          & 72.8          \\
    Rank-DETR \cite{pu2024rank}                & Swin-L   & 12     & 57.3          & 75.9          & 62.9          & 40.8          & 61.3          & 73.2          \\
    Rank-DINO \cite{pu2024rank}                & Swin-L   & 12     & 57.6          & 76.0          & \textbf{63.4} & \textbf{41.6} & 61.4          & 73.8          \\
    Relation-DETR                              & Swin-L   & 12     & \textbf{57.8} & \textbf{76.1} & 62.9          & 41.2          & \textbf{62.1} & \textbf{74.4} \\
    \bottomrule
  \end{tabular}
\end{table}

\subsubsection{Comparison on task-specific datasets.}
Different from generic object detection benchmarks, datasets in task-specific scenarios lack sufficient samples to provide semantic information. To reveal the generalizability of Relation-DETR, we conduct a performance comparison on two defect detection datasets, \ie CSD \cite{wang2023surface} and MSSD \cite{chen2023eee}. The results in \cref{tab:comparison on csd} show that, Relation-DETR improves the baseline DINO by 1.4\% AP on CSD, achieving the highest 54.4\% AP. \cref{tab:comparison on mssd} demonstrates that Relation-DETR further increases the margin to 6.4\% AP on MSSD and surpasses other counterparts. It is noteworthy that CSD \cite{wang2023surface} and MSSD \cite{chen2023eee} contain more small-sized objects than COCO 2017, which thus confirms the effectiveness of Relation-DETR to small-sized detection. Moreover, considering a stricter IoU threshold, Relation-DETR outperforms the second best method DINO by a significant margin of 11.1\% AP@75, highlighting the beneficial impact of explicit position relation on high-quality predictions.

\begin{table}[!htbp]
  \centering
  \caption{Quantitative comparison on CSD \cite{wang2023surface}.}
  \label{tab:comparison on csd}
  \begin{tabular}{@{}llccccccc@{}}
    \toprule
    Method                            & Backbone  & Epochs & AP@50:95      & AP@50         & AP@75         & AP$_S$        & AP$_M$        & AP$_L$        \\ \midrule
    Def-DETR \cite{zhu2020deformable} & ResNet-50 & 300    & 43.7          & 86.2          & 36.6          & 40.5          & 34.9          & \textbf{10.0} \\
    DAB-Def-DETR \cite{liu2021dab}    & ResNet-50 & 90     & 52.9          & 91.2          & 55.0          & 50.3          & 39.4          & 0.0           \\
    DN-Def-DETR \cite{li2022dn}       & ResNet-50 & 60     & 49.9          & 88.0          & 51.2          & 47.6          & 37.7          & 0.0           \\
    DINO \cite{zhang2022dino}         & ResNet-50 & 60     & 53.0          & 90.8          & 55.5          & 50.9          & 39.6          & 0.0           \\
    H-Def-DETR \cite{jia2023detrs}    & ResNet-50 & 60     & 53.0          & 90.6          & 55.7          & 51.2          & 39.2          & 6.7           \\
    Salience-DETR\cite{hou2024salience} & ResNet-50 & 60 & 53.2 & 92.5 & 55.1 & 51.0 & 40.9 & 0.0 \\
    Relation-DETR                     & ResNet-50 & 60     & \textbf{54.4} & \textbf{92.9} & \textbf{56.0} & \textbf{53.6} & \textbf{40.8} & 0.0           \\ \bottomrule
  \end{tabular}
\end{table}
\begin{table}[!htbp]
  \centering
  \caption{Quantitative comparison on MSSD \cite{chen2023eee}.}
  \label{tab:comparison on mssd}
  \begin{tabular}{@{}llccccccc@{}}
    \toprule
    Method                            & Backbone  & Epochs & AP@50:95 & AP@50 & AP@75 & AP$_S$ & AP$_M$ & AP$_L$ \\ \midrule
    Def-DETR \cite{zhu2020deformable} & ResNet-50 & 300    & 33.0     & 54.3  & 32.0  & 9.8    & 11.1   & 33.3   \\
    DAB-Def-DETR \cite{liu2021dab}    & ResNet-50 & 120    & 33.7     & 60.0  & 31.0  & 15.9   & 26.7   & 29.0   \\
    DN-Def-DETR \cite{li2022dn}       & ResNet-50 & 120    & 45.6     & 74.1  & 44.9  & 18.6   & 31.9   & 41.0   \\
    H-Def-DETR \cite{jia2023detrs}    & ResNet-50 & 120    & 46.9     & 76.8  & 47.1  & 20.3   & 45.3   & 40.7   \\
    DINO \cite{zhang2022dino}         & ResNet-50 & 120    & 51.0     & 80.0  & 52.5  & 20.0   & 47.4   & 44.8   \\
    Salience-DETR \cite{hou2024salience} & ResNet-50 & 120 & 55.4     & 78.2  & 61.9  & 28.7   & 47.5   & 44.5  \\
    Relation-DETR                     & ResNet-50 & 120    & \textbf{57.4}     & \textbf{79.2}  & \textbf{63.6}  & \textbf{31.4}   & \textbf{53.5}   & \textbf{45.5}   \\ \bottomrule
  \end{tabular}
\end{table}

\subsection{Ablation study}
This part conducts an ablation study to explore how the proposed components influence the final detection performance on COCO. The results in \cref{tab:abla} show that, each key component of Relation-DETR consistently contributes to improving AP. Even on a highly-optimized baseline with VariFocal loss \cite{zhang2021varifocalnet}, our position relation encoder and contrast pipeline bring clear improvements of +0.3\%, +0.5\% AP, respectively.
Built upon normalized relative geometry features, the position relation overcomes scale bias effectively, thus benefiting consistent performance improvements for different sized objects. For instance, \cref{tab:abla} shows that introducing relation into the baseline with VFL achieves +1.2\% AP$_S$, +1.0\% AP$_M$, and +1.3\% AP$_L$.

\begin{table}[htbp]
  \caption{Ablation study on key components of Relation-DETR (ResNet-50, 1x). We study both baseline and an optimized version with VariFocal Loss \cite{zhang2021varifocalnet}.}
  \label{tab:abla}
  \begin{tabular}{@{}ccccccccc@{}}
    \toprule
    VFL        & Rel.       & contrast   & AP@50:95                                       & AP@50                                          & AP@75                                          & AP$_S$                                         & AP$_M$                                          & AP$_L$                                         \\ \midrule
               &            &            & 49.9                                           & 67.4                                           & 54.5                                           & 33.9                                           & 53.5                                            & 63.8                                           \\
               & \checkmark &            & 50.3\textcolor{blue}{\textbf{($\uparrow$0.4)}} & 68.1\textcolor{blue}{\textbf{($\uparrow$0.7)}} & 55.0\textcolor{blue}{\textbf{($\uparrow$0.5)}} & 34.7\textcolor{blue}{\textbf{($\uparrow$0.8)}} & 53.4\textcolor{red}{\textbf{($\downarrow$0.1)}} & 64.7\textcolor{blue}{\textbf{($\uparrow$0.9)}} \\
    \checkmark &            &            & 50.9\textcolor{blue}{\textbf{($\uparrow$0.6)}} & 68.2\textcolor{blue}{\textbf{($\uparrow$0.1)}} & 55.3\textcolor{blue}{\textbf{($\uparrow$0.3)}} & 34.9\textcolor{blue}{\textbf{($\uparrow$0.2)}} & 54.6\textcolor{blue}{\textbf{($\uparrow$1.2)}}  & 64.8\textcolor{blue}{\textbf{($\uparrow$0.1)}} \\
    \checkmark & \checkmark &            & 51.2\textcolor{blue}{\textbf{($\uparrow$0.3)}} & 68.5\textcolor{blue}{\textbf{($\uparrow$0.7)}} & 55.7\textcolor{blue}{\textbf{($\uparrow$0.4)}} & 35.3\textcolor{blue}{\textbf{($\uparrow$0.4)}} & 54.9\textcolor{blue}{\textbf{($\uparrow$0.3)}}  & 65.5\textcolor{blue}{\textbf{($\uparrow$0.7)}} \\
    \checkmark & \checkmark & \checkmark & 51.7\textcolor{blue}{\textbf{($\uparrow$0.5)}} & 69.1\textcolor{blue}{\textbf{($\uparrow$0.6)}} & 56.3\textcolor{blue}{\textbf{($\uparrow$0.6)}} & 36.1\textcolor{blue}{\textbf{($\uparrow$0.8)}} & 55.6\textcolor{blue}{\textbf{($\uparrow$0.7)}}  & 66.1\textcolor{blue}{\textbf{($\uparrow$0.6)}} \\ \bottomrule
  \end{tabular}
\end{table}

\subsection{Transferability of position relation}
Our position relation encoder adopts an elegant architectural design, ensuring a promising transferability to existing DETR detectors with minimal modifications.
The experimental results in \cref{tab:transfer exp} show that, integrating position relation encoders without any further modifications enhances detection performance with clear margins of 1.6\%, 2.0\%, 0.1\% and 0.2\% for Deformable-DETR \cite{zhu2020deformable}, DAB-Deformable-DETR \cite{liu2021dab}, DN-Deformable-DETR \cite{li2022dn} and DINO \cite{zhang2022dino}, respectively. Interestingly, in comparison to AP$_M$ and AP$_L$, the position relation has a more substantial impact on improving AP$_S$ for Deformable-DETR \cite{zhu2020deformable} and DAB-Deformable-DETR \cite{liu2021dab}. We attribute this to the fact that these early proposed baselines introduce relatively less structural bias and thus benefit more from our explicit position relation prior.

\begin{table}[htbp]
  \caption{Transfer experiments for the position relation encoder (ResNet-50, 1$\times$). ``+RelEnc'' denotes the version that integrates our position relation encoder.}
  \label{tab:transfer exp}
  \begin{tabular}{@{}lcccccc@{}}
    \toprule
    Method       & AP@50:95                                       & AP@50                                          & AP@75                                          & AP$_S$                                          & AP$_M$                                          & AP$_L$                                          \\ \midrule
    Def-DETR     & 45.4                                           & 65.0                                           & 49.1                                           & 27.2                                            & 49.6                                            & 61.0                                            \\
    +RelEnc    & 47.0\textcolor{blue}{\textbf{($\uparrow$1.6)}} & 65.6\textcolor{blue}{\textbf{($\uparrow$0.6)}} & 51.2\textcolor{blue}{\textbf{($\uparrow$2.1)}} & 29.3\textcolor{blue}{\textbf{($\uparrow$2.1)}}  & 51.0\textcolor{blue}{\textbf{($\uparrow$1.4)}}  & 62.2\textcolor{blue}{\textbf{($\uparrow$1.2)}}  \\ \midrule
    DAB-Def-DETR & 46.3                                           & 65.9                                           & 50.4                                           & 28.3                                            & 50.5                                            & 61.9                                            \\
    +RelEnc    & 48.3\textcolor{blue}{\textbf{($\uparrow$2.0)}} & 66.5\textcolor{blue}{\textbf{($\uparrow$0.6)}} & 52.9\textcolor{blue}{\textbf{($\uparrow$2.5)}} & 32.4\textcolor{blue}{\textbf{($\uparrow$4.1)}}  & 52.0\textcolor{blue}{\textbf{($\uparrow$1.5)}}  & 62.0\textcolor{blue}{\textbf{($\uparrow$0.1)}}  \\ \midrule
    DN-Def-DETR  & 47.2                                           & 65.3                                           & 51.3                                           & 30.2                                            & 50.6                                            & 62.4                                            \\
    +RelEnc    & 47.3\textcolor{blue}{\textbf{($\uparrow$0.1)}} & 65.6\textcolor{blue}{\textbf{($\uparrow$0.3)}} & 51.4\textcolor{blue}{\textbf{($\uparrow$0.1)}} & 29.9\textcolor{red}{\textbf{($\downarrow$0.3)}} & 50.8\textcolor{blue}{\textbf{($\uparrow$0.2)}}  & 62.1\textcolor{red}{\textbf{($\downarrow$0.3)}} \\ \bottomrule
  \end{tabular}
\end{table}

Moreover, the proposed contrast pipeline can be seen as an extension of hybrid matching \cite{jia2023detrs}, utilizing the proposed position relation encoder. \Cref{tab:transfer hybrid} compares their transferability when integrated into DINO. The results indicate that directly applying hybrid matching \cite{jia2023detrs} to DINO \cite{zhang2022dino} results in a decrease in performance, from 49.9\% AP to 49.5\% AP. In contrast, the introduction of both the proposed relation encoder and the extended contrast pipeline consistently increases the performance. This demonstrates the effectiveness of the proposed position relation prior in improve detection performance, overcoming the weak generalizability inherent in hybrid matching.

\begin{table}
  \caption{Transfer experiments compared with hybrid matching \cite{jia2023detrs} on DINO.}
  \label{tab:transfer hybrid}
  \centering
  \setlength{\tabcolsep}{2.0pt}
  \begin{tabular}{@{}ccc|ccc|ccc|ccc@{}}
  \toprule
  \multicolumn{3}{c|}{DINO} & \multicolumn{3}{c|}{+Hybrid\cite{jia2023detrs}} & \multicolumn{3}{c|}{+RelEnc}      & \multicolumn{3}{c}{+RelEnc+Contrast} \\ \midrule
  AP     & AP@50   & AP@75  & AP       & AP@50     & AP@75     & AP       & AP@50     & AP@75     & AP          & AP@50        & AP@75       \\
  49.9   & 67.4    & 54.5   & 49.5$^\dagger$     & 66.6      & 54.0      & 50.3     & 68.1      & 55.0      & 51.2        & 68.7         & 55.6        \\ \bottomrule
  \end{tabular}
  \begin{tablenotes}
    \item \scriptsize $^\dagger$ We found that hybrid matching decreases the performance of DINO. The conclusion is consistent with the results of HDINO (see \href{https://github.com/open-mmlab/mmdetection/tree/main/projects/HDINO}{here}) reported by MMDetection\cite{chen2019mmdetection}.
  \end{tablenotes}
\end{table}

\subsection{Intuitive performance comparison}
\begin{figure}[htbp]
  \centering
  \includegraphics[width=0.48\textwidth]{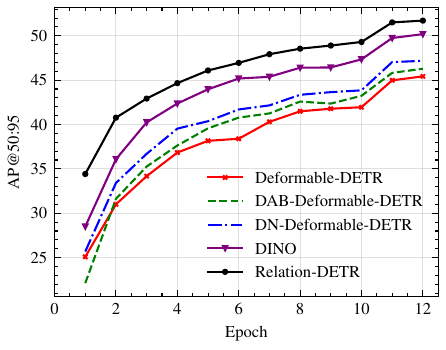}
  \includegraphics[width=0.48\textwidth]{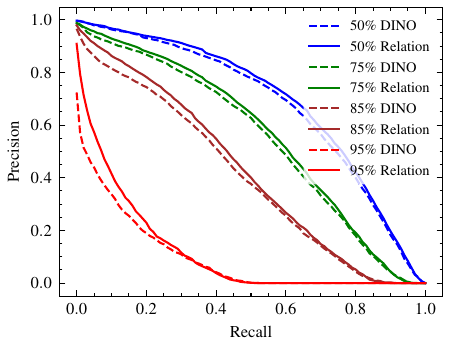}
  \caption{Convergence curve(left) and Precision-recall curve for IoU=$50\%\sim95\%$(right). All models are trained with ResNet-50 backbone under the same 1$\times$ training configuration on COCO 2017.}
  \label{fig:curve comparison}
\end{figure}

To facilitate an intuitive performance comparison, \cref{fig:curve comparison} plots the convergence curve and the precision-recall curve. Because the position relation prior reduces the requirement for learning structural bias from data \cite{lin2022survey}, Relation-DETR illustrates a faster convergence speed. When training from scratch, it achieves a higher AP than other counterparts with fewer iterations. Specifically, Relation-DETR can achieve over 40\% AP with only \textbf{2} epochs, surpassing existing DETR detectors. In addition to convergence speed, the PR curves under different IoU thresholds also validate the performance gain of our Relation-DETR.

\subsection{Visualization}
\label{sec:visualization}
\begin{figure}[htbp]
  \centering
  \begin{minipage}{0.58\textwidth}
    \centering
    \includegraphics[width=\textwidth]{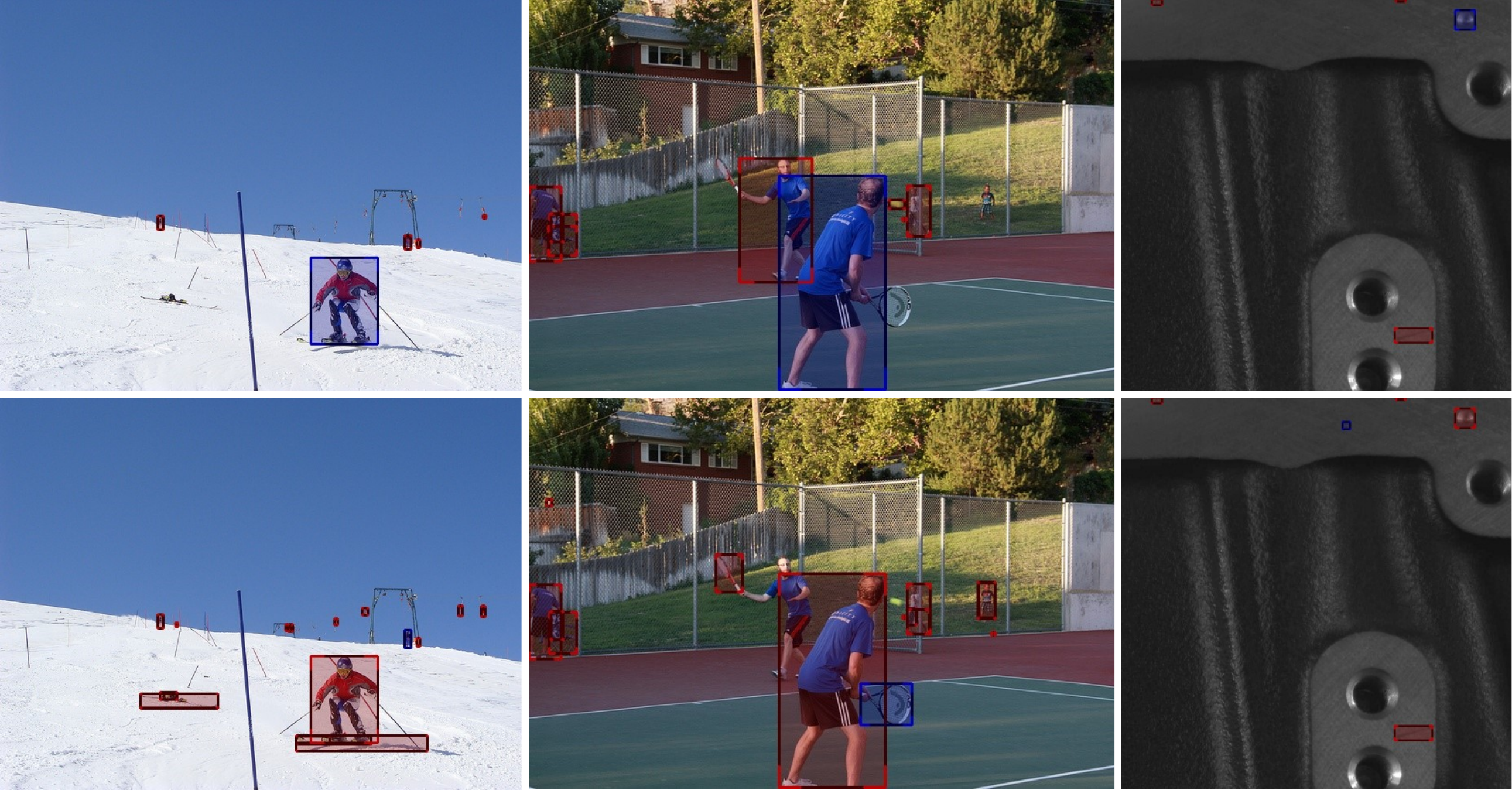}
    \caption{Representative objects(\textcolor{red}{red}) related to the given object(\textcolor{blue}{blue})}
    \label{fig:relation}

  \end{minipage}
  \begin{minipage}{0.405\textwidth}
    \centering
    \includegraphics[width=\textwidth]{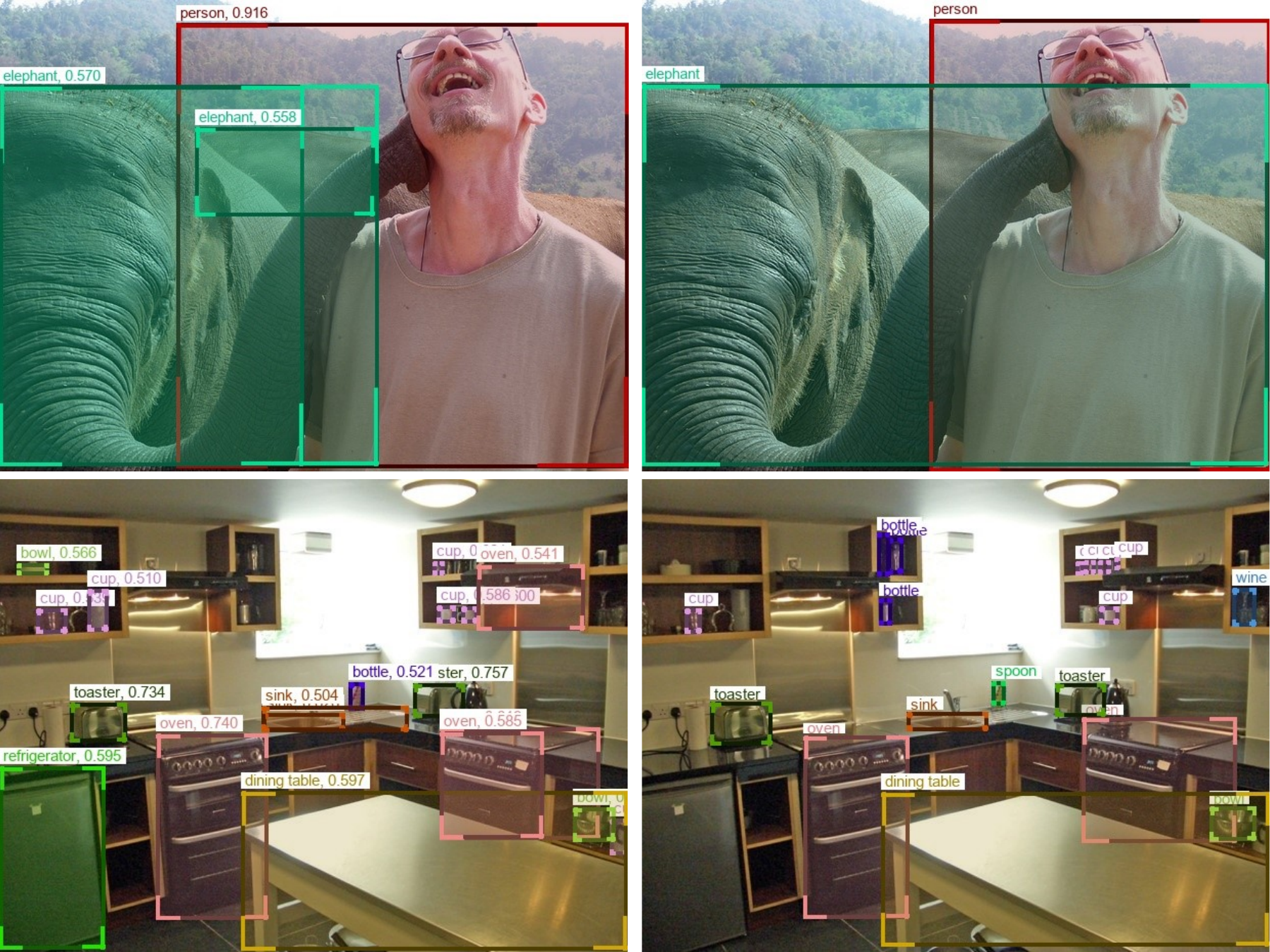}
    \caption{Predictions (left) and GTs (right) of failure cases}
    \label{fig:failure cases}
  \end{minipage}

\end{figure}

For a more intuitive grasp of the relation mechanism, \cref{fig:relation} illustrates  representative objects with high relation weights when given a query object. The visualization shows that for both generic and task-specific datasets, the relation contributes to identifying other detection candidates based on the given object query. Furthermore, small-sized objects tend to establish more relation connections with other objects due to the lack of their own semantic information. Therefore, constructing relations is crucial for small-sized object detection. \cref{fig:failure cases} further visualizes some failure cases of Relation-DETR, indicating that the presented model may benefit from occluded objects and dense objects with misleading semantic differences by considering more complex relations, such as occlusion and semantic relations.

\subsection{Towards universal object detection}
Will the position relation prior still be effective for datasets that cover a broader range of scenarios and objects? As a piece of general prior knowledge, we anticipate that the explicit position relation prior can benefit universal object detection tasks. To explore this, we have constructed a large-scale class-agnostic detection dataset with about 100,000 images, termed SA-Det-100k, by sampling a subset from SA-1B, which is one of the largest scale segmentation datasets proposed in Segment Anything \cite{kirillov2023segment}. We then compared the performance of our Relation-DETR with the baseline DINO \cite{zhang2022dino} using VFL \cite{zhang2021varifocalnet} on this dataset. The results in \cref{tab:universal} show that Relation-DETR achieves a clear improvement of 1.3\% AP, demonstrating the scalability of the proposed position relation prior.

\begin{table}[]
  \centering
  \caption{Quantitative comparison on SA-Det-100k (ResNet-50, 1$\times$).}
  \label{tab:universal}
  \begin{tabular}{@{}c|cccccc@{}}
  \toprule
  Methods       & AP@50:95 & AP@50 & AP@75 & AP$_S$ & AP$_M$ & AP$_L$ \\ \midrule
  DINO with VFL & 43.7     & 52.0  & 47.7  & 5.8 & 43.0 & 61.5 \\
  Relation-DETR & 45.0\textcolor{blue}{\textbf{($\uparrow$1.3)}}     & 53.1\textcolor{blue}{\textbf{($\uparrow$1.1)}}   & 48.9\textcolor{blue}{\textbf{($\uparrow$1.2)}}  & 6.0\textcolor{blue}{\textbf{($\uparrow$0.2)}} & 44.4\textcolor{blue}{\textbf{($\uparrow$1.4)}} & 62.9\textcolor{blue}{\textbf{($\uparrow$1.4)}} \\ \bottomrule
  \end{tabular}
\end{table}

\section{Conclusion}
\label{sec:conclu}
This paper explores explicit position relation prior for enhancing performance and convergence of DETR detectors. Built upon normalized relative geometry features, we propose a novel position relation that overcomes scale bias for progressive attention refinement. To address the conflicts between non-duplicate predictions and sufficient positive supervision in DETR frameworks, we extend the streaming pipeline to a contrast pipeline based on the proposed position relation.
Combining these components produces
a state-of-the-art detector, named Relation-DETR. Massive ablation studies and experimental results demonstrate the superior performance, faster convergence and promising transferability of the proposed detector. Moreover, Relation-DETR exhibits remarkable generalizability for both generic and task-specific detection tasks. We believe that this work will inspire future research on relation and structural bias for DETR detectors.

\section*{Acknowledgements}
This work was supported by the National Natural Science Foundation of China under Grant 62327808, and the Fundamental Research Funds for Xi'an Jiaotong University under Grants xtr072022001 and xzy022024009.

\bibliographystyle{splncs04}
\bibliography{main}

\begin{thebibliography}{10}
\providecommand{\url}[1]{\texttt{#1}}
\providecommand{\urlprefix}{URL }
\providecommand{\doi}[1]{https://doi.org/#1}

\bibitem{bodla2017soft}
Bodla, N., Singh, B., Chellappa, R., Davis, L.S.: {Soft-NMS}--improving object
  detection with one line of code. In: Int. Conf. Comput. Vis. pp. 5561--5569
  (2017)

\bibitem{bolya2019yolact}
Bolya, D., Zhou, C., Xiao, F., Lee, Y.J.: {YOLACT}: Real-time instance
  segmentation. In: Int. Conf. Comput. Vis. pp. 9157--9166 (2019)

\bibitem{cai2023align}
Cai, Z., Liu, S., Wang, G., Ge, Z., Zhang, X., Huang, D.: {Align-DETR}:
  Improving {DETR} with simple {IoU}-aware {BCE} loss. arXiv preprint
  arXiv:2304.07527  (2023)

\bibitem{carion2020end}
Carion, N., Massa, F., Synnaeve, G., Usunier, N., Kirillov, A., Zagoruyko, S.:
  End-to-end object detection with transformers. In: Eur. Conf. Comput. Vis.
  pp. 213--229. Springer (2020)

\bibitem{chen2019mmdetection}
Chen, K., Wang, J., Pang, J., Cao, Y., Xiong, Y., Li, X., Sun, S., Feng, W.,
  Liu, Z., Xu, J., et~al.: Mmdetection: Open mmlab detection toolbox and
  benchmark. arXiv preprint arXiv:1906.07155  (2019)

\bibitem{chen2023group}
Chen, Q., Chen, X., Wang, J., Zhang, S., Yao, K., Feng, H., Han, J., Ding, E.,
  Zeng, G., Wang, J.: {Group DETR}: Fast {DETR} training with group-wise
  one-to-many assignment. In: Int. Conf. Comput. Vis. pp. 6633--6642 (2023)

\bibitem{chen2018iterative}
Chen, X., Li, L.J., Fei-Fei, L., Gupta, A.: Iterative visual reasoning beyond
  convolutions. In: IEEE Conf. Comput. Vis. Pattern Recog. pp. 7239--7248
  (2018)

\bibitem{chen2023eee}
Chen, Y., Pan, J., Lei, J., Zeng, D., Wu, Z., Chen, C.: {EEE-Net}: Efficient
  edge enhanced network for surface defect detection of glass. IEEE
  Transactions on Instrumentation and Measurement  (2023)

\bibitem{cordts2016cityscapes}
Cordts, M., Omran, M., Ramos, S., Rehfeld, T., Enzweiler, M., Benenson, R.,
  Franke, U., Roth, S., Schiele, B.: The cityscapes dataset for semantic urban
  scene understanding. In: IEEE Conf. Comput. Vis. Pattern Recog. pp.
  3213--3223 (2016)

\bibitem{dai2021dynamic}
Dai, X., Chen, Y., Yang, J., Zhang, P., Yuan, L., Zhang, L.: {Dynamic DETR}:
  End-to-end object detection with dynamic attention. In: Int. Conf. Comput.
  Vis. pp. 2988--2997 (2021)

\bibitem{deng2009imagenet}
Deng, J., Dong, W., Socher, R., Li, L.J., Li, K., Fei-Fei, L.: {ImageNet}: A
  large-scale hierarchical image database. In: IEEE Conf. Comput. Vis. Pattern
  Recog. pp. 248--255. Ieee (2009)

\bibitem{dosovitskiy2020image}
Dosovitskiy, A., Beyer, L., Kolesnikov, A., Weissenborn, D., Zhai, X.,
  Unterthiner, T., Dehghani, M., Minderer, M., Heigold, G., Gelly, S., et~al.:
  An image is worth 16x16 words: Transformers for image recognition at scale.
  In: Int. Conf. Learn. Represent. (2020)

\bibitem{everingham2010pascal}
Everingham, M., Van~Gool, L., Williams, C.K., Winn, J., Zisserman, A.: The
  pascal visual object classes (voc) challenge. Int. J. Comput. Vis.
  \textbf{88},  303--338 (2010)

\bibitem{fang2023eva}
Fang, Y., Sun, Q., Wang, X., Huang, T., Wang, X., Cao, Y.: {EVA-02}: A visual
  representation for neon genesis. arXiv preprint arXiv:2303.11331  (2023)

\bibitem{feng2021tood}
Feng, C., Zhong, Y., Gao, Y., Scott, M.R., Huang, W.: {TOOD}: Task-aligned
  one-stage object detection. In: Int. Conf. Comput. Vis. pp. 3490--3499. IEEE
  Computer Society (2021)

\bibitem{girshick2015fast}
Girshick, R.: {Fast R-CNN}. In: Int. Conf. Comput. Vis. pp. 1440--1448 (2015)

\bibitem{hao2023relation}
Hao, X., Huang, D., Lin, J., Lin, C.Y.: Relation-enhanced {DETR} for component
  detection in graphic design reverse engineering. In: IJCAI. pp. 4785--4793
  (2023)

\bibitem{he2016deep}
He, K., Zhang, X., Ren, S., Sun, J.: Deep residual learning for image
  recognition. In: IEEE Conf. Comput. Vis. Pattern Recog. pp. 770--778 (2016)

\bibitem{hou2023canet}
Hou, X., Liu, M., Zhang, S., Wei, P., Chen, B.: {CANet}: Contextual information
  and spatial attention based network for detecting small defects in
  manufacturing industry. Pattern Recognition  \textbf{140},  109558 (2023)

\bibitem{hou2024salience}
Hou, X., Liu, M., Zhang, S., Wei, P., Chen, B.: Salience detr: Enhancing
  detection transformer with hierarchical salience filtering refinement. In:
  IEEE Conf. Comput. Vis. Pattern Recog. pp. 17574--17583 (2024)

\bibitem{hu2018relation}
Hu, H., Gu, J., Zhang, Z., Dai, J., Wei, Y.: Relation networks for object
  detection. In: IEEE Conf. Comput. Vis. Pattern Recog. pp. 3588--3597 (2018)

\bibitem{hu2024dac}
Hu, Z., Sun, Y., Wang, J., Yang, Y.: {DAC-DETR}: Divide the attention layers
  and conquer. Adv. Neural Inform. Process. Syst.  \textbf{36} (2024)

\bibitem{jia2023detrs}
Jia, D., Yuan, Y., He, H., Wu, X., Yu, H., Lin, W., Sun, L., Zhang, C., Hu, H.:
  {DETRs} with hybrid matching. In: IEEE Conf. Comput. Vis. Pattern Recog. pp.
  19702--19712 (2023)

\bibitem{jiang2018hybrid}
Jiang, C., Xu, H., Liang, X., Lin, L.: Hybrid knowledge routed modules for
  large-scale object detection. Adv. Neural Inform. Process. Syst.  \textbf{31}
  (2018)

\bibitem{kirillov2023segment}
Kirillov, A., Mintun, E., Ravi, N., Mao, H., Rolland, C., Gustafson, L., Xiao,
  T., Whitehead, S., Berg, A.C., Lo, W.Y., et~al.: Segment anything. In: IEEE
  Conf. Comput. Vis. Pattern Recog. pp. 4015--4026 (2023)

\bibitem{kolve2017ai2}
Kolve, E., Mottaghi, R., Han, W., VanderBilt, E., Weihs, L., Herrasti, A.,
  Deitke, M., Ehsani, K., Gordon, D., Zhu, Y., et~al.: Ai2-thor: An interactive
  3d environment for visual ai. arXiv preprint arXiv:1712.05474  (2017)

\bibitem{krishna2017visual}
Krishna, R., Zhu, Y., Groth, O., Johnson, J., Hata, K., Kravitz, J., Chen, S.,
  Kalantidis, Y., Li, L.J., Shamma, D.A., et~al.: Visual genome: Connecting
  language and vision using crowdsourced dense image annotations. Int. J.
  Comput. Vis.  \textbf{123},  32--73 (2017)

\bibitem{li2022dn}
Li, F., Zhang, H., Liu, S., Guo, J., Ni, L.M., Zhang, L.: {DN-DETR}: Accelerate
  {DETR} training by introducing query denoising. In: IEEE Conf. Comput. Vis.
  Pattern Recog. pp. 13619--13627 (2022)

\bibitem{li2020generalized}
Li, X., Wang, W., Wu, L., Chen, S., Hu, X., Li, J., Tang, J., Yang, J.:
  Generalized focal loss: Learning qualified and distributed bounding boxes for
  dense object detection. Adv. Neural Inform. Process. Syst.  \textbf{33},
  21002--21012 (2020)

\bibitem{li2022exploring}
Li, Y., Mao, H., Girshick, R., He, K.: Exploring plain vision transformer
  backbones for object detection. In: Eur. Conf. Comput. Vis. pp. 280--296.
  Springer (2022)

\bibitem{li2020gar}
Li, Z., Du, X., Cao, Y.: Gar: Graph assisted reasoning for object detection.
  In: Proceedings of the IEEE/CVF Winter Conference on Applications of Computer
  Vision. pp. 1295--1304 (2020)

\bibitem{lin2021core}
Lin, J., Pan, Y., Lai, R., Yang, X., Chao, H., Yao, T.: {Core-Text}: Improving
  scene text detection with contrastive relational reasoning. In: Int. Conf.
  Multimedia and Expo. pp.~1--6. IEEE (2021)

\bibitem{lin2022survey}
Lin, T., Wang, Y., Liu, X., Qiu, X.: A survey of transformers. AI Open  (2022)

\bibitem{lin2017focal}
Lin, T.Y., Goyal, P., Girshick, R., He, K., Doll{\'a}r, P.: Focal loss for
  dense object detection. In: Int. Conf. Comput. Vis. pp. 2980--2988 (2017)

\bibitem{lin2014microsoft}
Lin, T.Y., Maire, M., Belongie, S., Hays, J., Perona, P., Ramanan, D.,
  Doll{\'a}r, P., Zitnick, C.L.: Microsoft {COCO}: Common objects in context.
  In: Eur. Conf. Comput. Vis. pp. 740--755. Springer (2014)

\bibitem{liu2021dab}
Liu, S., Li, F., Zhang, H., Yang, X., Qi, X., Su, H., Zhu, J., Zhang, L.:
  {DAB-DETR}: Dynamic anchor boxes are better queries for {DETR}. In: Int.
  Conf. Learn. Represent. (2021)

\bibitem{liu2023detection}
Liu, S., Ren, T., Chen, J., Zeng, Z., Zhang, H., Li, F., Li, H., Huang, J., Su,
  H., Zhu, J., et~al.: Detection transformer with stable matching. arXiv
  preprint arXiv:2304.04742  (2023)

\bibitem{liu2019adaptive}
Liu, S., Huang, D., Wang, Y.: {Adaptive NNS}: Refining pedestrian detection in
  a crowd. In: IEEE Conf. Comput. Vis. Pattern Recog. pp. 6459--6468 (2019)

\bibitem{liu2021swin}
Liu, Z., Lin, Y., Cao, Y., Hu, H., Wei, Y., Zhang, Z., Lin, S., Guo, B.: Swin
  transformer: Hierarchical vision transformer using shifted windows. In: Int.
  Conf. Comput. Vis. pp. 10012--10022 (2021)

\bibitem{pu2024rank}
Pu, Y., Liang, W., Hao, Y., Yuan, Y., Yang, Y., Zhang, C., Hu, H., Huang, G.:
  {Rank-DETR} for high quality object detection. Adv. Neural Inform. Process.
  Syst.  \textbf{36} (2024)

\bibitem{shao2019objects365}
Shao, S., Li, Z., Zhang, T., Peng, C., Yu, G., Zhang, X., Li, J., Sun, J.:
  Objects365: A large-scale, high-quality dataset for object detection. In:
  Int. Conf. Comput. Vis. pp. 8430--8439 (2019)

\bibitem{vaswani2017attention}
Vaswani, A., Shazeer, N., Parmar, N., Uszkoreit, J., Jones, L., Gomez, A.N.,
  Kaiser, {\L}., Polosukhin, I.: Attention is all you need. Adv. Neural Inform.
  Process. Syst.  \textbf{30} (2017)

\bibitem{wang2023surface}
Wang, Q., Gao, S., Xiong, L., Liang, A., Jiang, K., Zhang, W.: A casting
  surface dataset and benchmark for subtle and confusable defect detection in
  complex contexts. IEEE Sensors Journal pp.~1--1 (2024)

\bibitem{wang2022anchor}
Wang, Y., Zhang, X., Yang, T., Sun, J.: {Anchor DETR}: Query design for
  transformer-based detector. In: AAAI. vol.~36, pp. 2567--2575 (2022)

\bibitem{xu2019reasoning}
Xu, H., Jiang, C., Liang, X., Lin, L., Li, Z.: {Reasoning-RCNN}: Unifying
  adaptive global reasoning into large-scale object detection. In: IEEE Conf.
  Comput. Vis. Pattern Recog. pp. 6419--6428 (2019)

\bibitem{ye2023cascade}
Ye, M., Ke, L., Li, S., Tai, Y.W., Tang, C.K., Danelljan, M., Yu, F.:
  {Cascade-DETR}: Delving into high-quality universal object detection. In:
  Int. Conf. Comput. Vis. pp. 6704--6714 (2023)

\bibitem{zhang2022dino}
Zhang, H., Li, F., Liu, S., Zhang, L., Su, H., Zhu, J., Ni, L., Shum, H.Y.:
  {DINO}: {DETR} with improved denoising anchor boxes for end-to-end object
  detection. In: Int. Conf. Learn. Represent. (2022)

\bibitem{zhang2021varifocalnet}
Zhang, H., Wang, Y., Dayoub, F., Sunderhauf, N.: Varifocalnet: An iou-aware
  dense object detector. In: IEEE Conf. Comput. Vis. Pattern Recog. pp.
  8514--8523 (2021)

\bibitem{zhang2023dense}
Zhang, S., Wang, X., Wang, J., Pang, J., Lyu, C., Zhang, W., Luo, P., Chen, K.:
  Dense distinct query for end-to-end object detection. In: IEEE Conf. Comput.
  Vis. Pattern Recog. pp. 7329--7338 (2023)

\bibitem{zhao2024ms}
Zhao, C., Sun, Y., Wang, W., Chen, Q., Ding, E., Yang, Y., Wang, J.: Ms-detr:
  Efficient detr training with mixed supervision. In: IEEE Conf. Comput. Vis.
  Pattern Recog. pp. 17027--17036 (2024)

\bibitem{zhu2020deformable}
Zhu, X., Su, W., Lu, L., Li, B., Wang, X., Dai, J.: {Deformable DETR}:
  Deformable transformers for end-to-end object detection. In: Int. Conf.
  Learn. Represent. (2020)

\bibitem{zong2023detrs}
Zong, Z., Song, G., Liu, Y.: {DETRs} with collaborative hybrid assignments
  training. In: Int. Conf. Comput. Vis. pp. 6748--6758 (2023)

\end{thebibliography}
\end{document}